\documentclass[letterpaper, 10 pt, conference]{ieeeconf}  
\IEEEoverridecommandlockouts                              
\usepackage{url}
\usepackage{xcolor}
\usepackage{booktabs}
\usepackage{float}
\usepackage{amssymb}
\usepackage{colortbl}
\usepackage{cite}
\usepackage{graphicx}
\usepackage{amsmath}
\usepackage{dsfont}
\usepackage{mathtools}
\usepackage{amssymb}
\DeclareMathOperator*{\argmin}{arg\,min}
\usepackage{algorithm}
\usepackage{algorithmicx}
\usepackage[noend]{algcompatible}
\usepackage{algpseudocode}
\usepackage{algcompatible}
\usepackage{booktabs}
\usepackage{multirow}
\usepackage{subcaption}
\usepackage[table]{xcolor}

\title{\LARGE \bf
{Back to the Future: Predicting Long-Horizon Dynamics with Retrospective Correction for Disturbance-Aware Retrieval
}
}

\author{Jiwoong Choi$^{1}$, Haoyu Li$^{2}$, and Yiwei Lyu$^{2}$%
\thanks{\scriptsize $^{1}$Dept. of Aerospace Eng. and Eng. Mechanics, University of Texas at Austin, $^{2}$Dept. of Computer Sci. and Eng., Texas A\&M University}%
}

\begin{document}

\maketitle
\thispagestyle{empty}
\pagestyle{empty}

\begin{abstract}
Retrieving objects from densely packed storage requires anticipating how contact disturbs surrounding objects when limited clearance prevents lifting the target over clutter. However, contact can propagate beyond the target's immediate neighbors, while accumulated prediction errors make learned dynamics unreliable over long planning horizons. We introduce DARE (Disturbance-Aware REtrieval), a geometry-aware relational dynamics framework that uses retrospective correction to improve long-horizon prediction for disturbance-aware retrieval. DARE encodes compact object descriptors and predicts action-conditioned disturbance propagation through object-level message passing. When a sparse observation becomes available, a learned corrector uses the prediction discrepancy to amend an earlier predicted state and reruns the intervening dynamics, propagating the correction through the predicted contact chain. These predictions guide a receding-horizon planner that balances retrieval progress against displacement and rotation of surrounding objects, permitting necessary contact while discouraging harmful disturbance. Experiments in planar and RoboCasa environments evaluate long-horizon prediction, generalization, and robustness for disturbance-aware retrieval in dense clutter. See our project website: \url{the-back-to-the-future.github.io}.
\end{abstract}

\section{Introduction}

Humans effortlessly retrieve a wide variety of objects around cluttered environment everyday, such as picking up eggs from refrigerator, retrieving plates from dishwasher and grasping tableware from cabinet. Retrieving target objects from clutter environment is not merely a grasping problem~\cite{danielczuk2019mechanical}. In densely packed and vertically constrained storage, such as refrigerator, cabinet or dishwasher, limited clearance and height preclude lifting target object over the clutter. Retrieval therefore requires intentional contact with neighboring objects or environment. Moreover, forces imparted by the moving target propagate through successive object to object contacts, producing disturbance cascades in which directly and indirectly affected objects slide, rotate, topple, or fall. Consequently, a geometrically feasible trajectory remains physically unacceptable if it spills or damages objects. 

Prior works overcome this by rearranging the environment and building a feasible trajectory for retrieving target objects~\cite{danielczuk2019mechanical, huangmechanical, liu2025fetchbot, chen2025clutterdexgrasp, bai2025retrdex}. However, these approaches primarily achieve feasibility by circumventing clutter and provide limited reasoning about the physical consequences of unavoidable contact, limiting long-horizon planning, for example, pulling a bottle past a neighboring can risks pushing the can into a glass near the shelf edge, toppling the glass without direct bottle--glass contact. Learned dynamics models offer a practical alternative by explicitly forecasting how candidate target motions affect surrounding objects, enabling physics-aware planning that balances retrieval progress against collateral disturbance~\cite{fignet, pfaff2021mgn, rubanova2024learning, wei2025integrating}. While recent advances show promising results leveraging Graph Neural Network (GNN) to approximating real-world contacts~\cite{battaglia2016interaction, li2019propagation}, the accumulated drift error prohibit long-horizon motion prediction under shifts in object geometry and scene composition.

\begin{figure}[t]
  \centering
  \includegraphics[width=0.95\columnwidth]{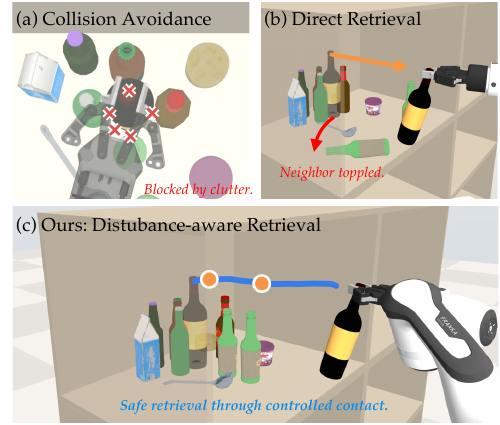}
  \caption{\textbf{Disturbance-aware retrieval.} (a) Collision avoidance leaves the target blocked. (b) Direct retrieval topples a neighboring object. (c) Our method predicts contact-induced disturbance to plan retrieval through controlled contact, preserving stability of surrounding objects.}
  \label{fig:teaser}
  \vspace{-6mm}
\end{figure}

To address these challenges, we introduce \textbf{DARE}, a relational dynamics learning framework with retrospective correction for disturbance-aware object retrieval. The framework encodes object geometry into compact descriptors and uses an object-level Propagation Network to predict how candidate target motions induce multi-object disturbance cascades. When a sparse observation becomes available, a learned error attributor uses the observed discrepancy to amend earlier predicted states and regenerate the subsequent rollout. We jointly train the dynamics model and corrector through this prediction--correction cycle, with a multi-step objective extending beyond the observation time to promote corrections that improve future predictions. The resulting model guides a closed-loop model predictive path integral (MPPI) planner that balances retrieval progress against disturbance to surrounding objects. To summary, \textbf{our key contributions} are: 1) A geometry-aware relational dynamics model that combines compact mesh-derived representations with action-conditioned message passing to predict multi-object disturbance cascades across diverse object shapes and scene sizes; 2) A retrospective correction module that uses sparse observations to revise earlier predicted states and propagate their corrections forward, addressing accumulated error in long-horizon dynamics prediction; and 3) We perform extensive analysis and ablation study across three scenarios to provide insights into DARE's learning procedure and performance gains.

\section{Related Work}

\begin{figure*}[t]
    \centering
    \includegraphics[width=\textwidth]{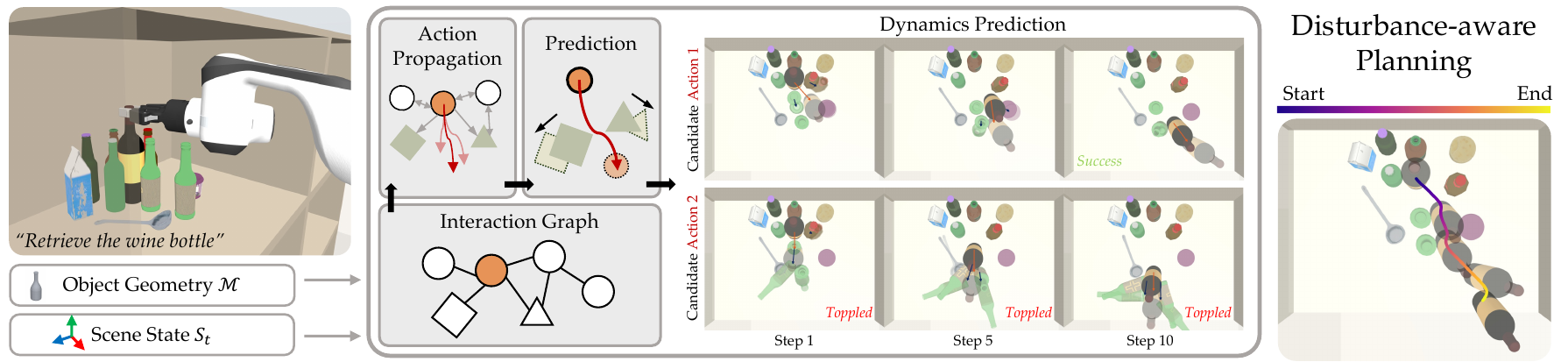}
    \caption{\textbf{Pipeline Overview:} Given object geometry and the current scene state, \textbf{DARE} propagates action-conditioned interactions over an object graph to predict scene evolution under candidate target motions. The resulting rollouts capture direct and indirect disturbances to surrounding objects, enabling the planner to select a retrieval trajectory that balances goal progress against collateral disturbance.}
    \label{fig:pipeline}
    \vspace{-5mm}
\end{figure*}

\noindent\textit{\textbf{Object Retrieval in Cluttered Environment.}} Retrieving target objects from cluttered environment has gathered significant attention in robotic manipulation. For this task, some prior works~\cite{danielczuk2019mechanical, huangmechanical, liu2025fetchbot, chen2025clutterdexgrasp, bai2025retrdex} rearrange or navigate the clutter to expose the target and establish an extraction path.~\cite{danielczuk2019mechanical} iteratively selects pushing, suction and grasping actions to remove occluders. A learning-based method generates minimal disturbance trajectory from depth observations to grasp target objects~\cite{liu2025fetchbot}. Another line of works explicitly predict physical consequences of candidate actions, considering simultaneous object contacts~\cite{Dogar-RSS-12}, multi-object transition~\cite{huang2021visual} and local dynamics propagation~\cite{Li-RSS-24} to infer feasible retrieval pose or safe removal order. Other works leverage force and tactile feedback to adjust their policies during execution to ensure safe contact~\cite{brouwer2025gentle}. Despite these advances, existing retrieval pipelines provide limited support for assessing the long-horizon consequences of cascading contacts. DARE addresses this gap by forecasting multi-object disturbances and revising these predictions as observations arrive, enabling retrieval planning that jointly considers extraction feasibility and collateral disturbance.

\noindent\textit{\textbf{Relational Dynamics Learning.}} Modeling the dynamics of objects is challenging due to the high complexity of the state spaces and the variance of physical properties. Traditional physics engines~\cite{todorov2012mujoco} predict motion by numerically solving hand-crafted optimization problems. Learning-based methods represent scenes as interacting entities and infer their dynamics from interaction data. Interaction Network~\cite{battaglia2016interaction} introduces object- and relation-centric graph that learns pairwise interactions. Propagation Network~\cite{li2019propagation} proposes multi-step message passing to capture indirect effects across multiple objects under partial observation. To model rigid objects pioneering works rely on mesh-based inputs to faithfully capture the complex dynamics~\cite{fignet, pfaff2021mgn, rubanova2024learning, wei2025integrating}. FIGNet~\cite{fignet} constructs interaction over mesh faces to improve collision modeling. HOPNet~\cite{wei2025integrating} incorporates higher-order topology and physics-informed pathways to improve long-horizon propagation. Although mesh-level interaction models capture detailed contact geometry, their message-passing cost grows with geometric resolution. DARE instead encodes each object's collision geometry into a compact per-object descriptor and propagates action-conditioned interactions at object-level, supporting diverse shapes and variable object counts.

\noindent\textit{\textbf{State Estimation and Correction in Learned Dynamics Models.}} Existing methods mitigate simulation drift through learned residual correction~\cite{um2020solver}, learned constraint enforcement~\cite{yang2020learning}, and denoising-based prediction refinement~\cite{lippe2023pde}. Observation-conditioned state updates incorporate incoming measurements through Kalman filtering~\cite{kalman1960new} or graph-based sensor fusion, while learned trajectory smoothing propagates later evidence into earlier estimates through backward recursion or hybrid message passing~\cite{revach2023rtsnet}. Complementarily, simulator-based inverse estimation reconstructs dynamics by optimizing initial conditions or physical parameters against sparse observations~\cite{zhao2022learning}. Our approach learns retrospective state corrections under a multi-step prediction objective that extends beyond the corrected observation interval, explicitly training each amendment to improve subsequent predictions.

\section{Method}
\begin{algorithm}[t]\footnotesize
    \caption{One cycle of disturbance-constrained retrieval}
    \label{alg:retrieval}
    \begin{algorithmic}[1]
    \Require scene estimate $\tilde{\mathbf{s}}_0$, opening $(\mathbf{p}_m,\mathbf{n}_m)$, dynamics model $f_{\theta}$, corrector $h_{\xi}$, nominal knots $\bar{\boldsymbol{\kappa}}$
    \Ensure the next $\tilde{\mathbf{s}}_0$ and $\bar{\boldsymbol{\kappa}}$
    \For{each round}
        \State sample $\mathcal{U}$ around $\bar{\boldsymbol{\kappa}}$ through~\eqref{eq:spline_plan}, clamped to $\mathcal{A}$
        \State roll every $\mathbf{u}\in\mathcal{U}$ out with $f_{\theta}$ from $\tilde{\mathbf{s}}_0$
        \State keep $\mathcal{U}\cap\mathcal{S}_{\mathrm{safe}}$~\eqref{eq:safe_set}; score it with~\eqref{eq:planning_cost}
        \State set $\bar{\boldsymbol{\kappa}}$ to the cost-weighted mean of the admissible knots
    \EndFor
    \State $\mathbf{u}^{*}\gets$ lowest-cost admissible candidate~\eqref{eq:executed_plan}, or hold still
    \State execute $\mathbf{a}_{0:H_{\mathrm{e}}-1}$ of $\mathbf{u}^{*}$; observe $\mathbf{o}_W$
    \State $\hat{\mathbf{s}}_{1:W}\gets$ blind rollout of the executed actions~\eqref{eq:blind_rollout}
    \State $\tilde{\mathbf{s}}_1\gets$ amend $\hat{\mathbf{s}}_1$ with $h_{\xi}$~\eqref{eq:retrospective_corrector}--\eqref{eq:corrected_orientation}
    \State $\tilde{\mathbf{s}}_W\gets$ reroll from $\tilde{\mathbf{s}}_1$~\eqref{eq:corrected_rerollout}
    \State $\bar{\boldsymbol{\kappa}}\gets\mathbf{u}^{*}$ advanced by $H_{\mathrm{e}}$ steps, refit to the knots
    \State \Return $\tilde{\mathbf{s}}_0\gets\tilde{\mathbf{s}}_W$, $\bar{\boldsymbol{\kappa}}$
    \end{algorithmic}
    
\end{algorithm}

\vspace{-2pt}
In this section, we first present the overall problem formulation (\S~\ref{sec:problem_formulation}). We then discuss how our relational dynamics model predicts the propagation of object interactions (\S~\ref{sec:dynamics_prediction}), how retrospective correction refines subsequent predictions (\S~\ref{sec:retro_correction} and \S~\ref{sec:joint_training}), and how these predictions guide closed-loop retrieval planning (\S~\ref{sec:planning}). Algorithm~\ref{alg:retrieval} summarizes the complete pipeline.

\subsection{Problem Statement}
\label{sec:problem_formulation}

\noindent\textbf{\textit{Setup.}}
We consider retrieving a grasped target ($i=0$) from a fixed, densely cluttered compartment $\mathcal{B}$ containing $N$ bystanders ($i=1,\ldots,N$). Limited overhead clearance prevents lifting above neighbors, so retrieval requires contact that propagates displacement, toppling, or ejection. Each object is described by collision geometry $\mathcal{G}_i$, mass $m_i$, inertia $I_i$, and state $\mathbf{s}_{i,t}=(\mathbf{p}_{i,t},\mathbf{q}_{i,t},\mathbf{v}_{i,t},\boldsymbol{\omega}_{i,t})$, comprising position, orientation, and linear and angular velocities. The scene state is $\mathbf{s}_t=\{\mathbf{s}_{i,t}\}_{i=0}^{N}$. The robot controls the target's planar translation, tilt, and yaw through rate commands $\mathbf{a}_t\in\mathcal{A}$. The action set $\mathcal{A}$ imposes fixed bounds on each rate, while a separate constraint limits the target's tilt angle. The target follows these commands, whereas bystanders move through contact interactions:
\begin{align}
    \mathbf{s}_{0,t+1}
    &=\mathcal{T}(\mathbf{s}_{0,t},\mathbf{a}_t),
    \nonumber\\
    \mathbf{s}_{i,t+1}
    &=f\!\left(\mathbf{s}_t,\mathbf{a}_t;
    \{\mathcal{G}_j,m_j,I_j\}_{j},\mathcal{B}\right),
    \quad i=1,\ldots,N,
    \label{eq:dynamics_formulation}
\end{align}
where $\mathcal{T}$ integrates the commanded rates over one step and $f$ describes the unknown contact dynamics of the bystanders. We approximate $f$ with a learned relational model $f_\theta$ to predict the consequences of candidate target motions. Sparse, noisy observations provide feedback to correct errors accumulated during autoregressive prediction.

\noindent\textbf{\textit{Objective.}}
Given the current scene estimate $\tilde{\mathbf{s}}_0$, we plan an $H$-step sequence $\mathbf{u}=(\mathbf{a}_0,\ldots,\mathbf{a}_{H-1})\in\mathcal{A}^{H}$ by solving
\begin{equation}
    \min_{\mathbf{u}\in\mathcal{S}_{\mathrm{safe}}}
    J_{\mathrm{prog}}(\mathbf{u})+J_{\mathrm{dist}}(\mathbf{u}),
    \label{eq:planning_formulation}
\end{equation}
where $\mathcal{S}_{\mathrm{safe}}\subseteq\mathcal{A}^{H}$ enforces predicted terminal displacement and per-step tilt tolerances for every bystander. $J_{\mathrm{prog}}$ measures the target's signed, capped terminal center-of-mass distance from the opening, and $J_{\mathrm{dist}}$ penalizes all bystanders' terminal displacement, tilt, and yaw. After executing the first $H_{\mathrm{e}}$ actions, the robot observes and replans.

\subsection{Relational Dynamics Prediction}
\label{sec:dynamics_prediction}
Planning in Sec.~\ref{sec:problem_formulation} requires efficient prediction of candidate motions' consequences. Contact responses depend on geometry beyond bounding boxes (footprints, silhouettes, mass distributions), and a target disturbs objects indirectly as forces propagate through intermediate ones. We therefore adopt an object-level relational model~\cite{li2019propagation,tekden2024pusheffect}: one node per rigid object, a learned collision-geometry descriptor, and repeated message passing, so that disturbance chains are predicted \textit{efficiently} across many candidate trajectories.

\noindent\textbf{\textit{Object Representation.}} Representing each object by scalar dimensions~\cite{tekden2024pusheffect} is efficient but
cannot adequately distinguish the diverse shapes encountered in
household environments. Conversely, constructing a graph over mesh
vertices or faces~\cite{fignet,wei2025integrating} preserves detailed geometry but makes inference cost particularly restrictive during planning, where the dynamics model evaluates many actions and future steps.

We thus compress each object's collision geometry into a
fixed-dimensional descriptor that is computed once and reused
throughout the rollout. Let $\mathcal{G}_i=\{g_{i,k}\}_{k=1}^{K_i}$
denote the convex decomposition of object $i$ with $K_i$ convex pieces.
Each convex piece is summarized by a feature vector $\mathbf{x}_{i,k}$ containing its centroid, volume, principal-axis orientation, and inertia-ellipsoid semi-axes.

We encode the unordered convex-piece features into a permutation-invariant object descriptor $\mathbf{z}_i$ leveraging~\cite{zaheer2017deepsets}. Joint training with the dynamics model encourages this representation to capture geometric features relevant to physical response prediction. However, geometry alone cannot fully represent contact dynamics. We therefore
combine $\mathbf{z}_i$ with the object's mass and inertia:
\begin{equation}
    \mathbf{c}_i
    =
    \left[
        \mathds{1}_{i=0},
        m_i,
        \operatorname{vec}(I_i),
        \mathbf{z}_i
    \right],
    \label{eq:physical_descriptor}
\end{equation}
where $\mathds{1}_{i=0}$ identifies the commanded target and $\operatorname{vec}(I_i)$ holds the six independent components of the inertia tensor. This design retains one node per object, making the graph size independent of the resolution of collision meshes.

\noindent\textbf{\textit{Propagation Network.}}
We adopt the propagation network of~\cite{li2019propagation} as the backbone: objects are nodes, nearby objects are linked by proximity edges~\cite{tekden2024pusheffect} rebuilt after every predicted step, $L$ rounds of residual message passing over a propagated latent state $\bar{\mathbf{h}}_{i,t}^{(\ell)}$ relay a push through intermediate objects to every node it affects, and a decoder $D_{\theta}$ maps $\bar{\mathbf{h}}_{i,t}^{(L)}$ to a body-frame velocity per object~\cite{tekden2024pusheffect}, which we take as the mean linear and angular rates $(\bar{\mathbf{v}}_{i,t}^{b},\bar{\boldsymbol{\omega}}_{i,t}^{b})$ over the coming step. Our additions are the physical content of the graph. Each closed face of the compartment is a static node, connected to nearby objects through the perpendicular offset from the face plane to the object's center, so object--boundary contact is handled by the same relational computation as object--object contact. The encoders read
\begin{align}
    \mathbf{h}_{i,t}^{(0)}
    &=
    E_v
    \left(
        \mathbf{v}_{i,t}^{b},
        \boldsymbol{\omega}_{i,t}^{b},
        \mathbf{R}_{i,t},
        \mathbf{c}_i
    \right),
    \label{eq:node_encoder}\\
    \mathbf{r}_{ji,t}
    &=
    E_e
    \left(
        \Delta\mathbf{p}_{ji,t}^{\,i},
        \Delta\mathbf{R}_{ji,t}^{\,i},
        \Delta\mathbf{v}_{ji,t}^{\,i},
        \Delta\boldsymbol{\omega}_{ji,t}^{\,i},
        \mathbf{c}_j,
        \mathbf{c}_i,
        \mathbf{a}_{ji,t}^{\,i}
    \right),
    \label{eq:edge_encoder}
\end{align}
where the superscript $b$ denotes a body-frame quantity, $\mathbf{R}_{i,t}$ is the orientation as the first two columns of its rotation matrix, and the relative pose and rates of the edge from sender $j$ to receiver $i$ are expressed in the receiver's frame~\cite{tekden2024pusheffect}. Every node also carries an exit flag that freezes an object once it has left the compartment and its CoM offset in world axes, which places its weight relative to its support as it tilts. The action feature $\mathbf{a}_{ji,t}^{\,i}$ is the target's commanded rate over the coming step, in the receiver's frame, on edges sent by the target ($j=0$) and zero elsewhere: the action reaches a bystander only through a relational path from the target, so the model must explain indirect motion through intermediate objects rather than broadcast the command to the scene. The decoded rates are integrated over a model step $\Delta t$ on $SE(3)$, the position by the rotated linear rate and the orientation by the quaternion exponential of the angular rate, while the target pose advances by its commanded kinematic motion~\eqref{eq:dynamics_formulation}. We next introduce a retrospective correction (\S~\ref{sec:retro_correction}) that uses sparse observations to repair the error that accumulates over autoregressive rollouts.

\subsection{Retrospective Correction from Sparse Observations}
\label{sec:retro_correction}

\begin{figure}
    \centering
    \includegraphics[width=.9\linewidth]{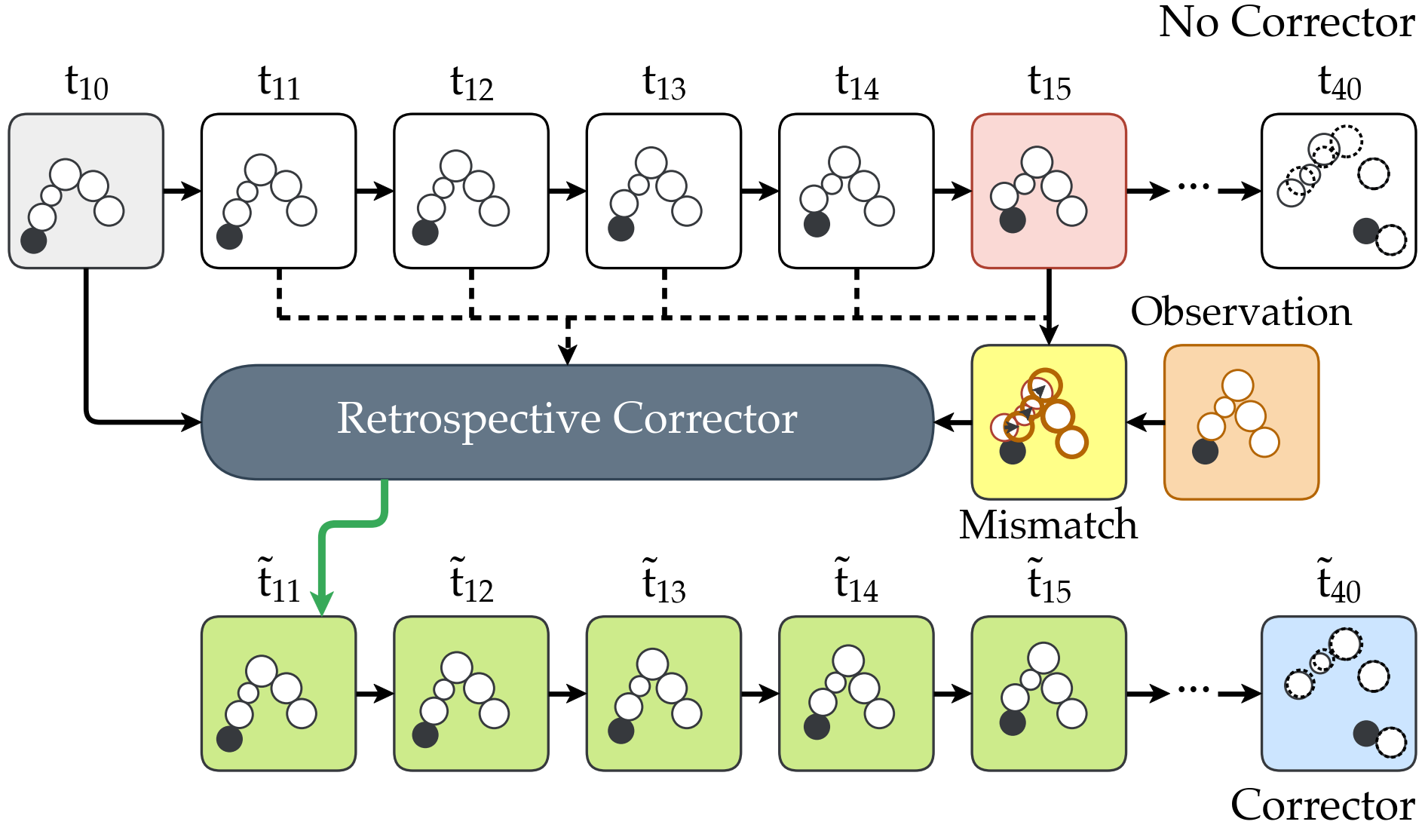}
    \caption{\textbf{Retrospective correction for long-horizon prediction.} The corrector combines prior predictions with the observation mismatch at $t_{15}$ (yellow) to estimate a corrected state $\tilde{t}_{11}$, the first step after the last observation $t_{10}$. Rerolling from it propagates the correction through the learned dynamics and initializes the next cycle, mitigating error \textit{drift}.}
    \label{fig:corrector}
    \vspace{-5mm}
\end{figure}

The relational model can predict multi-object interactions, but its
errors compound when predicted states are autoregressively fed back into the
model. This problem is especially severe in clutter: a small pose error
can shift the timing of a contact, alter the subsequent interaction
graph, and eventually produce a qualitatively different disturbance
cascade.

A fresh observation exposes this accumulated error, but not its cause. The simplest response is to overwrite the predicted state with the observation, which treats the observed pose as if it were a complete and dynamically consistent state~\cite{freeman2019learning}. An observation carries poses only, so the velocities of the reset state must be reconstructed from consecutive observations. When observations are \textit{sparse}, that finite difference averages over every contact inside the interval, leaving the reset state incomplete. When observations are \textit{noisy}, the sensor error is copied straight into the next rollout, so a simple overwriting can compound error rather than remove it. \textbf{DARE} instead changes an earlier state and reapplies the learned dynamics, so that the observation is weighed against what the dynamics can produce and the state carried forward is one the dynamics generated, with consistent velocities and contacts.

Consider an interval of $W$ prediction steps between two observations.
We use a local time index and denote the state carried from the previous
cycle by $\tilde{\mathbf{s}}_0$, which at the first cycle is the initial observation. From this state, the dynamics model
produces an autoregressive \textit{blind} rollout:
\begin{equation}
    \hat{\mathbf{s}}_{t+1}
    =
    f_{\theta}
    \left(
        \hat{\mathbf{s}}_t,
        \mathbf{a}_t
    \right),
    \qquad
    t=0,\ldots,W-1,
    \label{eq:blind_rollout}
\end{equation}
where $\hat{\mathbf{s}}_0=\tilde{\mathbf{s}}_0$ and the target follows its commanded motion~\eqref{eq:dynamics_formulation} at every step.

As the next observation $\mathbf{o}_W$ of every object's pose arrives, we compute the pose
discrepancy for each bystander $i$:
\begin{equation}
    \mathbf{e}^{\mathrm{o}}_{i,W}
    =
    \begin{bmatrix}
        \mathbf{p}^{\mathrm{o}}_{i,W}-\hat{\mathbf{p}}_{i,W}\\[1mm]
        \operatorname{Log}
        \left(
            \hat{\mathbf{q}}_{i,W}^{-1}
            \otimes
            \mathbf{q}^{\mathrm{o}}_{i,W}
        \right)
    \end{bmatrix},
    \label{eq:observation_error}
\end{equation}
where the superscript $\mathrm{o}$ marks the observed pose and $\operatorname{Log}(\cdot)$, the inverse of $\operatorname{Exp}(\cdot)$, maps the relative quaternion to a rotation vector, so that the error is signed as observed minus predicted. This discrepancy reflects where the rollout ends up
wrong, but still not how that error developed over the preceding interval.

To recover how the error developed, the retrospective corrector $h_{\xi}$, an MLP with parameters $\xi$, reads the endpoint discrepancy together with the carried state and every frame of the \textit{blind} rollout, $\hat{\mathbf{s}}_{1:W}$, that $f_{\theta}$ predicted before the observation arrived:
\begin{equation}
    \left\{
        \boldsymbol{\mu}_{i,t},
        \boldsymbol{\sigma}_{i,t}
    \right\}_{t=1}^{W}
    =
    h_{\xi}
    \left(
        \tilde{\mathbf{s}}_{i,0},
        \hat{\mathbf{s}}_{i,1:W},
        \mathbf{e}^{\mathrm{o}}_{i,W}
    \right).
    \label{eq:retrospective_corrector}
\end{equation}
For each frame, the output parameterizes a diagonal Gaussian over the
signed pose error. Predicting the error sequence provides dense
supervision for learning how an observed endpoint discrepancy relates
to error accumulated earlier in the rollout. Because the endpoint discrepancy is the compounded result of a deviation early in the interval rather than a spike of one-step error, the amendment is applied at the earliest frame, not at the frame where the error is largest. Reapplying the dynamics from there regenerates the whole interval, contacts included, so the correction propagates through the interaction history instead of patching its end. The carried state $\tilde{\mathbf{s}}_{0}$ anchors the cycle and is not amended, so frame~1 is the earliest state that can be. At inference time, we therefore use the predicted mean at the first rollout step to amend that state, for every bystander still in the compartment, at every observation:
\begin{align}
    \tilde{\mathbf{p}}_{i,1}
    &=
    \hat{\mathbf{p}}_{i,1}
    +
    \boldsymbol{\mu}^{p}_{i,1},
    \label{eq:corrected_position}\\
    \tilde{\mathbf{q}}_{i,1}
    &=
    \hat{\mathbf{q}}_{i,1}
    \otimes
    \operatorname{Exp}
    \left(
        \boldsymbol{\mu}^{q}_{i,1}
    \right),
    \label{eq:corrected_orientation}
\end{align}
where $\boldsymbol{\mu}^{p}_{i,1}$ and
$\boldsymbol{\mu}^{q}_{i,1}$ are the translational and rotational
components of the correction. The blind velocity of frame~1 describes the motion into $\hat{\mathbf{p}}_{i,1}$, not into $\tilde{\mathbf{p}}_{i,1}$, so we finite-difference it as the mean rate of the transition from $\tilde{\mathbf{s}}_{i,0}$ to the amended pose. Since the model relates consecutive poses through exactly this one-step mean rate, the amended frame remains a state the model was trained to consume. We then rerun the dynamics from it:
\begin{equation}
    \tilde{\mathbf{s}}_{t+1}
    =
    f_{\theta}
    \left(
        \tilde{\mathbf{s}}_t,
        \mathbf{a}_t
    \right),
    \qquad
    t=1,\ldots,W-1.
    \label{eq:corrected_rerollout}
\end{equation}
The rerolled state $\tilde{\mathbf{s}}_W$, rather than a direct
copy of $\mathbf{o}_W$, is carried into the next prediction cycle. During training, the reroll continues for $C$ further steps beyond the observation, which is what makes the amendment accountable for the future it produces (\S~\ref{sec:joint_training}).

\subsection{Future-aware Joint Predictor–Corrector Learning}
\label{sec:joint_training}

\begin{figure}[!t]
    \centering
    \includegraphics[width=.9\linewidth]{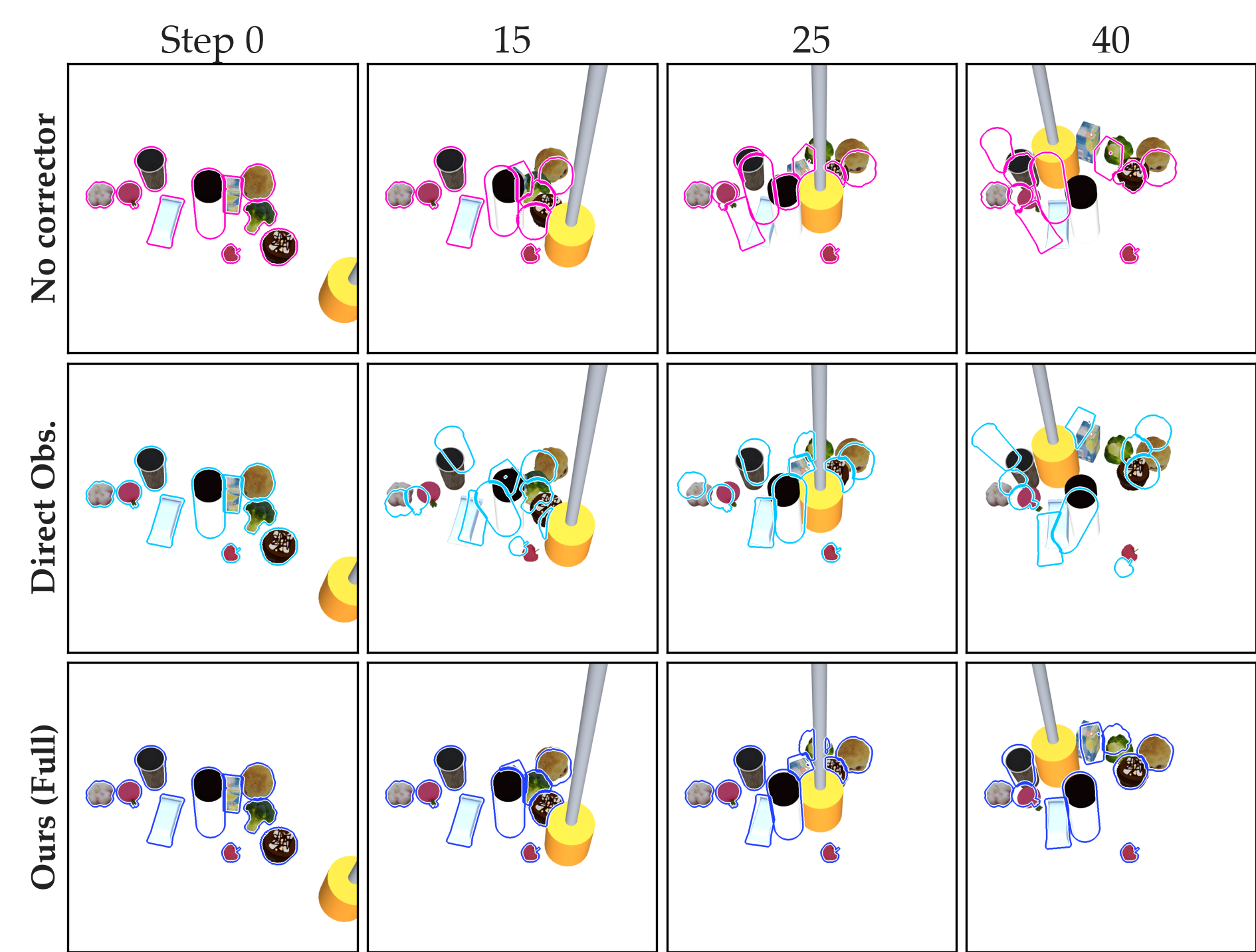}
    \caption{\textbf{Tabletop dynamics prediction under noise.} Outlines in purple, light blue and dark blue, indicate predicted object poses in a held-out scene as the target (yellow) pushes through clutter. Uncorrected predictions drift (top), direct observation updates inherit noise (middle), and noise-trained correction maintains alignment with the objects (bottom).}
    \label{fig:E2-fig-viz}
    \vspace{-5mm}
\end{figure}

Retrospective correction would reduce to trajectory fitting if it were
trained only to reproduce the observation at step $W$. Such a
correction may explain the available measurement while producing a
state from which the model predicts the subsequent dynamics poorly.
We therefore train the corrected rollout against a continuation of the
reference (ground-truth) trajectory beyond the observation.

Each training sample contains a reference trajectory
$\mathbf{s}^{*}_{0:W+C}$. The model first generates the blind rollout
$\hat{\mathbf{s}}_{1:W}$, after which the corrector produces an
amendment and the model rerolls to obtain
$\tilde{\mathbf{s}}_{1:W+C}$. Let
$\mathcal{D}(\mathbf{s}_{a:b},\mathbf{s}^{*}_{a:b})$ denote the trajectory error over frames $a$ to $b$: the root-mean-square position error plus $w_R$ times the root-mean-square rotation error, taken over every scored bystander frame, with the target, the boundary faces, and any bystander that has left the compartment removed from the mean rather than counted as zero error. We supervise the \textit{blind} rollout
with
\begin{equation}
    \mathcal{L}_{\mathrm{pre}}
    =
    \mathcal{D}\!\left(
        \hat{\mathbf{s}}_{1:W},
        \mathbf{s}^{*}_{1:W}
    \right).
    \label{eq:blind_loss}
\end{equation}
This term preserves the predictive accuracy of the underlying dynamics
model rather than relying on the corrector.

The \textit{corrected} rollout is supervised over both the observation interval
and the future continuation, as one root mean square over both ranges with equal weight per frame:

\begin{equation}
    \begin{aligned}
        \mathcal{L}_{\mathrm{post}}
        &=
        \mathcal{D}\!\left(
            \tilde{\mathbf{s}}_{1:W+C},
            \mathbf{s}^{*}_{1:W+C}
        \right),\\
        1:W{+}C
        &=
        \underbrace{1,\dots,W}_{\text{observed interval}}
        ,\;
        \underbrace{W{+}1,\dots,W{+}C}_{\textbf{continuation}}.
    \end{aligned}
    \label{eq:future_aware_loss}
\end{equation}
The continuation frames \textit{beyond the observation} are central to our formulation. An amendment that merely reproduces $\mathbf{o}_W$ fits frame $W$ at best, copies the observation's noise at worst, and in either case predicts poorly after it. In contrast, an amendment that repairs the cause of the error keeps predicting the reference trajectory for $C$ further steps. Scoring both ranges at the same weight therefore rewards an amendment only when the corrected state remains beneficial after the observation as well, thereby distinguishing future-aware correction from retrospective smoothing alone.

We also train the corrector to estimate the errors within the blind rollout, the per-frame Gaussian of~\eqref{eq:retrospective_corrector}. Let $\mathbf{e}^{*}_{i,t}$ be the signed pose error between
$\hat{\mathbf{s}}_{i,t}$ and the corresponding reference state. The
negative log-likelihood, averaged over the bystanders, is
\begin{equation}
    \mathcal{L}_{\mathrm{nll}}
    =
    \frac{1}{N}\sum_{i=1}^{N}\sum_{t=1}^{W}
    \left(
    \left\|
        \frac{\mathbf{e}^{*}_{i,t}-\boldsymbol{\mu}_{i,t}}{\sqrt{2}\,\boldsymbol{\sigma}_{i,t}}
    \right\|^{2}
    +
    \mathbf{1}^{\top}\log\boldsymbol{\sigma}_{i,t}
    \right),
    \label{eq:correction_nll}
\end{equation}
where the division and the logarithm are elementwise over the six pose channels and frames of a bystander that has left the compartment are excluded from the average. This term supervises the whole error sequence and keeps the predicted spread $\boldsymbol{\sigma}$ trained. The complete objective is
\begin{equation}
    \mathcal{L}
    =
    w_{\mathrm{post}}\mathcal{L}_{\mathrm{post}}
    +
    \lambda\mathcal{L}_{\mathrm{pre}}
    +
    \beta\mathcal{L}_{\mathrm{nll}}.
    \label{eq:joint_objective}
\end{equation}

The predictor and corrector are optimized \textit{jointly}: one backward pass runs through the complete blind-rollout, amendment, and rerollout procedure, with no gradient detached along the way. Consequently, the future-prediction loss shapes both modules: the corrector learns to propose dynamically useful amendments, while the predictor learns rollouts whose accumulated errors can be corrected from sparse observations. The predictor is thus trained to be correctable, not only accurate on its own.

\subsection{Disturbance-Aware Retrieval Planning}
\label{sec:planning}

The planner solves~\eqref{eq:planning_formulation} on the corrected rollouts: rather than rejecting all contact, it evaluates the predicted consequences of each candidate motion and excludes those likely to displace, topple, or eject bystanders. Each cycle starts from the scene estimate $\tilde{\mathbf{s}}_0$ carried by the correction cycle of \S~\ref{sec:retro_correction}; a candidate $\mathbf{u}\in\mathcal{A}^{H}$ is rolled out as in~\eqref{eq:dynamics_formulation} and~\eqref{eq:blind_rollout}, and $\hat{\mathbf{s}}_t(\mathbf{u})$ denotes the scene imagined $t$ steps into it.

\begin{figure}
    \centering
    \includegraphics[width=.9\linewidth]{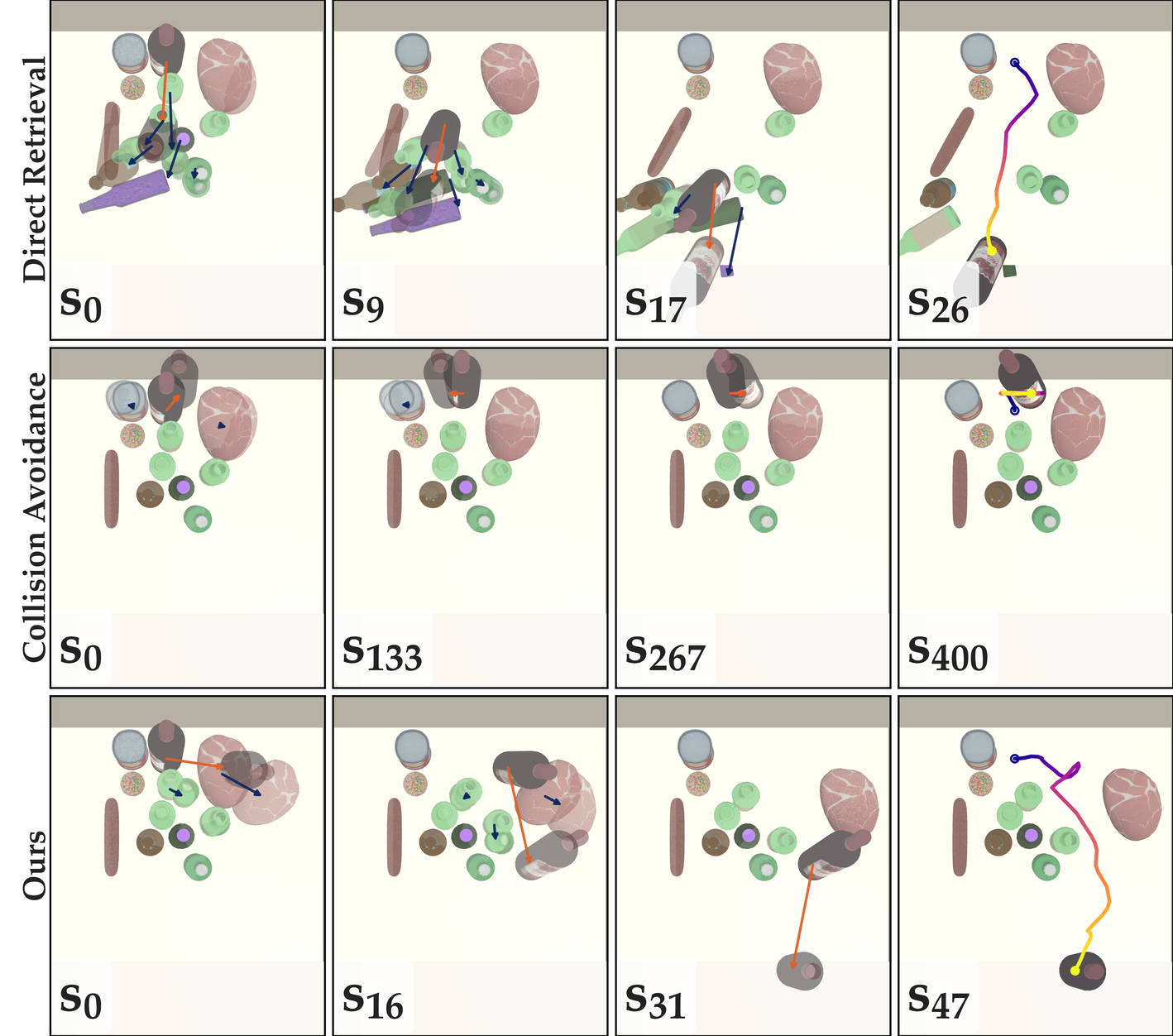}
    \caption{\textbf{Object Retrieval Visualization} Top-down views compare three planners, with target motion marked in orange. Each row shows four evenly spaced execution steps with translucent next-step poses and the executed target path in the final frame.}
    \label{fig:e3-3d-viz}
    \vspace{-5mm}
\end{figure}

\noindent\textit{\textbf{Objective.}}
With the opening defined by a point $\mathbf{p}_m$ on its plane and its horizontal outward unit normal $\mathbf{n}_m$, the target's distance to retrieval is the terminal shortfall of its CoM,
\begin{equation}
    d_{\mathrm{exit}}(\mathbf{u})
    =
    \mathbf{n}_m^{\top}
    \left(
        \mathbf{p}_m-\hat{\mathbf{p}}_{0,H}(\mathbf{u})
    \right),
    \label{eq:exit_shortfall}
\end{equation}
which becomes negative once the target crosses the opening. For each bystander we measure its terminal displacement $\Delta\mathbf{p}_{i,H}$, tilt $\tau_{i,H}$, and yaw $\psi_{i,H}$, each relative to its start pose. The cost of~\eqref{eq:planning_formulation} is then:
\begin{equation}
\begin{split}
    J(\mathbf{u})
    ={}&
    \underbrace{
    w_{\mathrm{term}}
    \max\!\left(d_{\mathrm{exit}}(\mathbf{u}),-d_{\max}\right)}_{J_{\mathrm{prog}}(\mathbf{u})}\\
    &+
    \underbrace{
    \sum_{i=1}^{N}
    \left(
        w_p\|\Delta\mathbf{p}_{i,H}\|_2
        +
        w_{\tau}\tau_{i,H}
        +
        w_{\psi}\psi_{i,H}
    \right)}_{J_{\mathrm{dist}}(\mathbf{u})}.
\end{split}
\label{eq:planning_cost}
\end{equation}
The credit cap $d_{\max}$ keeps the progress term from rewarding motion beyond the opening. Tilt and yaw carry separate weights so that tilt, which leads more directly to toppling, can be priced higher than yaw.

\noindent\textit{\textbf{Safety.}}
A candidate is admissible when all bystanders stay within the terminal displacement and per-step tilt tolerances:
\begin{equation}
    \mathcal{S}_{\mathrm{safe}}
    =
    \left\{
        \mathbf{u}\in\mathcal{A}^{H}
        :\;
        \|\Delta\mathbf{p}_{i,H}\|_2\leq\delta_p,\;
        \max_{t\leq H}\tau_{i,t}\leq\tau_{\max},
        \ \forall i
    \right\}.
    \label{eq:safe_set}
\end{equation}
Displacement is bounded only at the end of the plan, as objects may be nudged along the way to open a path. Tilt is bounded at every step, following the state-based failure condition of safe nudging~\cite{jin2026nudge}: tipping past a prescribed angle marks the onset of unsafe toppling. The rate bounds hold by construction and the target driver enforces its tilt threshold, so only bystanders are constrained.

\noindent\textbf{\textit{Ranking and selection.}}
We smooth the motion and reduce the sampling dimension by representing each plan as a cubic spline through a small set of knot values $\boldsymbol\kappa$,
\begin{equation}
    \mathbf{u}=\mathbf{B}\boldsymbol{\kappa},
    \label{eq:spline_plan}
\end{equation}
where $\mathbf{B}$ is the fixed spline interpolation matrix that maps the knots to the $H$ per-step actions. Each round perturbs a nominal knot sequence $\bar{\boldsymbol{\kappa}}$ into a candidate set $\mathcal{U}$, rolls every candidate out with the learned model, discards those outside $\mathcal{S}_{\mathrm{safe}}$, and replaces the nominal with the cost-weighted mean of the admissible knots under MPPI-style exponentiated weights~\cite{williams2018mppi}, with the cost shifted by the round's minimum and normalized by its admissible range so that the temperature is unit-free. After the last round the executed plan is the lowest-cost admissible candidate seen,
\begin{equation}
    \mathbf{u}^{*}
    =
    \argmin_{\mathbf{u}\in\mathcal{U}\cap\mathcal{S}_{\mathrm{safe}}}
    J(\mathbf{u}),
    \label{eq:executed_plan}
\end{equation}
rather than the nominal, as a plan the model actually imagined and scored is preferred to a blend of several. If no candidate is admissible in any round, the target holds still for the cycle rather than force progress.

\noindent\textbf{\textit{Execution and correction.}}
The executed prefix is the observation interval of \S~\ref{sec:retro_correction}, $H_{\mathrm{e}}=W$, so every planning cycle is one correction cycle and the rerolled $\tilde{\mathbf{s}}_W$ is the $\tilde{\mathbf{s}}_0$ of the next. The next solve is warm-started from the executed plan advanced by $H_{\mathrm{e}}$ steps, and the loop ends when the target's CoM passes the opening.

\section{Experiments}
\begin{figure}[!t]
    \centering
    \includegraphics[width=\linewidth]{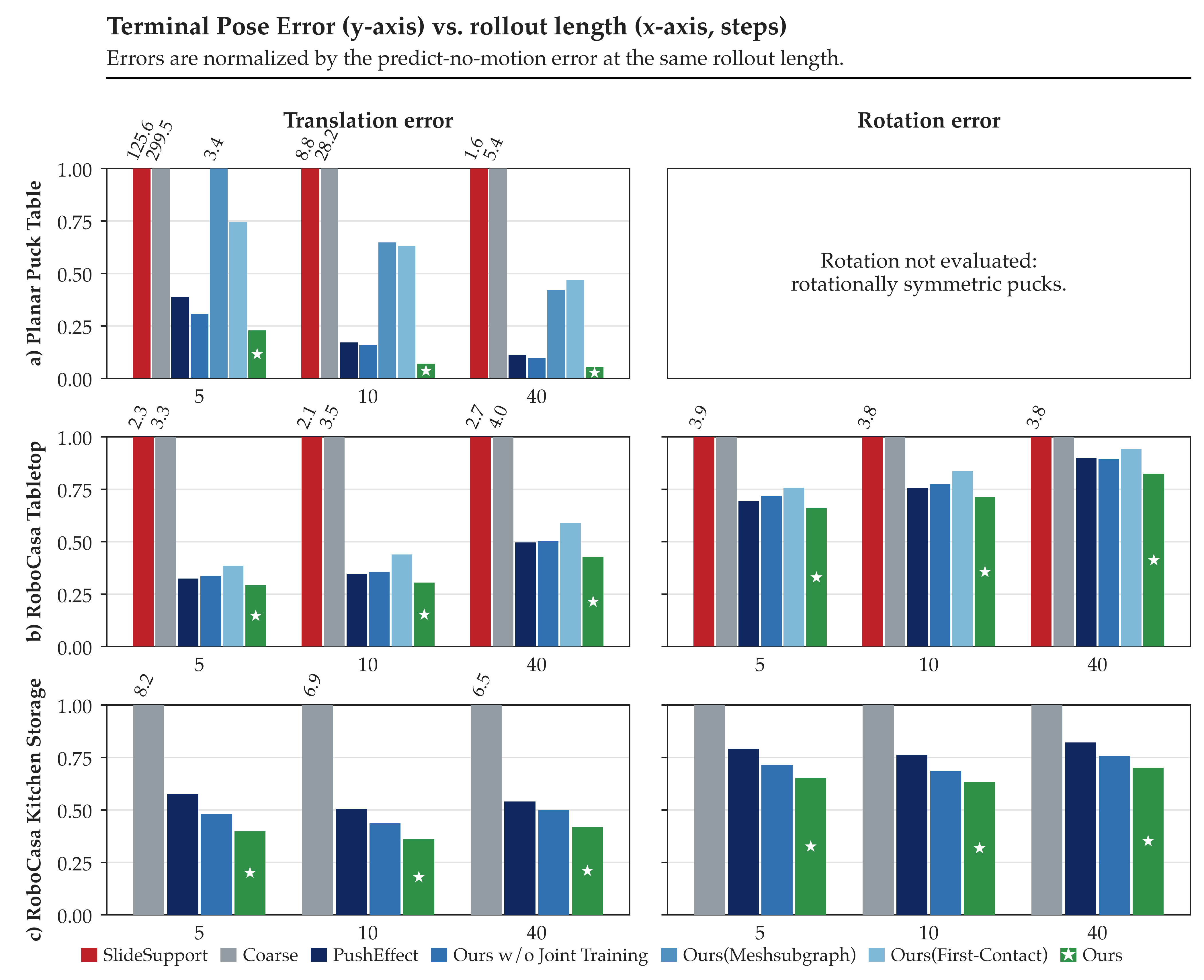}
    \caption{\textbf{Dynamics prediction.} Terminal translation (left) and rotation (right) errors normalized by the no-motion error at each horizon. Rotation is omitted for symmetric pucks.}
    \label{fig:E1-fig-1}
    \vspace{-5mm}
\end{figure}

\begin{table*}[t]
    \centering
    \caption{
        Retrieval from cluttered compartments. Each numerical entry reports Easy~/~Hard setup.
    }
    \label{tab:e3-retrieval-combined}

    \scriptsize
    \setlength{\tabcolsep}{2.2pt}
    \renewcommand{\arraystretch}{1.05}

    \resizebox{\textwidth}{!}{%
    \begin{tabular}{@{}llc*{12}{c}@{}}
        \toprule
        \multirow{2}{*}{Method}
        & \multirow{2}{*}{Obs.}
        & \multirow{2}{*}{Corr.}
        & \multicolumn{3}{c}{Success rate $\uparrow$}
        & \multicolumn{3}{c}{Translational disturbance (cm) $\downarrow$}
        & \multicolumn{3}{c}{Tilt disturbance ($^\circ$) $\downarrow$}
        & \multicolumn{3}{c}{Yaw disturbance ($^\circ$) $\downarrow$} \\
        \cmidrule(lr){4-6}
        \cmidrule(lr){7-9}
        \cmidrule(lr){10-12}
        \cmidrule(l){13-15}
        & & &
        5 obj. & 10 obj. & 20 obj. &
        5 obj. & 10 obj. & 20 obj. &
        5 obj. & 10 obj. & 20 obj. &
        5 obj. & 10 obj. & 20 obj. \\
        \midrule

        Direct retrieval
        & Clean & --
        & 0.60/0.60 & 0.30/0.20 & 0.20/0.10
        & 26.9/21.5 & 46.0/48.4 & 91.2/69.6
        & 66.5/36.5 & 120.3/88.9 & 191.6/105.2
        & 185.8/182.9 & 292.2/290.3 & 555.0/573.3 \\

        Collision avoidance
        & Clean & --
        & 0.40/0.50 & 0.20/0.10 & 0.10/0.10
        & 4.5/4.2 & 6.2/4.7 & 11.9/13.4
        & 3.3/24.3 & 20.8/17.7 & 54.2/70.9
        & 43.0/32.0 & 61.0/64.9 & 93.2/108.5 \\

        \rowcolor{gray!15}
        Ours w/o dist. obj.
        & Clean & Off
        & \textbf{0.90}/\textbf{1.00}
        & \textbf{0.90}/\textbf{1.00}
        & \textbf{0.90}/0.80
        & 11.1/9.6 & 35.1/28.4 & 81.7/84.0
        & 17.3/0.6 & 84.0/23.7 & 98.7/115.0
        & 127.7/142.7 & 286.8/198.1 & 486.9/538.1 \\

        \rowcolor{gray!15}
        Ours
        & Clean & Off
        & \textbf{0.90}/\textbf{1.00}
        & \textbf{0.90}/\textbf{1.00}
        & \textbf{0.90}/\textbf{1.00}
        & 9.9/8.5 & 30.0/33.3 & 75.1/71.2
        & 15.0/0.5 & 90.1/21.2 & 120.2/50.7
        & 113.6/126.1 & 300.5/223.2 & 458.0/500.4 \\

        \addlinespace[3pt]

        Ours w/o dist. obj.
        & Noisy & Off
        & 0.90/\textbf{1.00}
        & 0.40/0.00
        & 0.10/0.00
        & 11.1/11.6 & 28.0/14.3 & 87.7/77.5
        & 17.3/18.6 & 91.8/34.1 & 181.6/204.2
        & 119.3/148.9 & 240.6/78.7 & 458.2/513.3 \\

        Ours w/o dist. obj.
        & Noisy & On
        & \textbf{1.00}/\textbf{1.00}
        & \textbf{0.90}/\textbf{0.90}
        & \textbf{0.80}/\textbf{0.60}
        & 11.3/8.3 & 31.6/38.1 & 99.2/85.1
        & 35.5/9.8 & 53.4/16.2 & 151.5/96.8
        & 131.8/114.7 & 275.8/248.4 & 554.3/638.5 \\

        \rowcolor{gray!15}
        Ours
        & Noisy & Off
        & \textbf{1.00}/0.80
        & 0.00/0.00
        & 0.10/0.00
        & 12.2/9.5 & 13.1/4.3 & 57.2/61.5
        & 15.3/9.6 & 19.7/22.8 & 208.5/119.0
        & 129.8/104.6 & 98.3/30.4 & 400.1/397.9 \\

        \rowcolor{gray!15}
        Ours
        & Noisy & On
        & \textbf{1.00}/\textbf{1.00}
        & \textbf{1.00}/\textbf{0.80}
        & \textbf{0.30}/\textbf{0.30}
        & 7.1/8.2 & 30.9/32.1 & 97.7/83.1
        & 6.9/1.0 & 52.4/45.1 & 134.3/133.2
        & 94.4/118.2 & 238.6/246.7 & 568.7/507.8 \\

        \bottomrule
    \end{tabular}%
    }
    \vspace{-5mm}
\end{table*}

Our evaluation focuses on two questions: 1) How robust is our dynamics prediction to increasing scene complexity, longer rollout horizons, and unseen objects? 2) How does physical reasoning enable safe and feasible retrieval when contact is unavoidable in cluttered environments?

\subsection{Experiments Setup}

\noindent \textbf{\textit{Datasets.}} We evaluate on held-out episodes from three environments: a planar puck table, RoboCasa tabletop scenes, and RoboCasa kitchen cabinets~\cite{robocasa365}. The puck dataset contains 1,000 episodes, each with five pucks at randomized initial positions and a randomly sampled target goal. RoboCasa episodes vary compartment geometry, object composition and arrangement, and target motion. Each samples one of 13 compartments with varying floor dimensions and overhead clearance, along with height-compatible objects. To diversify contact configurations, we randomize the target's initial position and yaw and neighboring-object assignments, relative directions, gaps, and orientations. After physics settling, we move the target for 90 seconds through random planar waypoints with temporally correlated yaw and tilt perturbations. These variations produce diverse contact chains and object responses, including displacement and potential falls.

\noindent \textbf{Metrics.}
For dynamics prediction, we report final-timestep translation and rotation RMSE. For object retrieval, we report success rate and collateral disturbance, measured by non-target objects' translational and rotational displacements from their initial to post-retrieval states. Success requires the target to reach its goal along a feasible trajectory without any object tilting by $90^\circ$ or falling out of the compartment.

\subsection{Dynamics Prediction}

\begin{figure}[!t]
    \centering
    \includegraphics[width=1\linewidth]{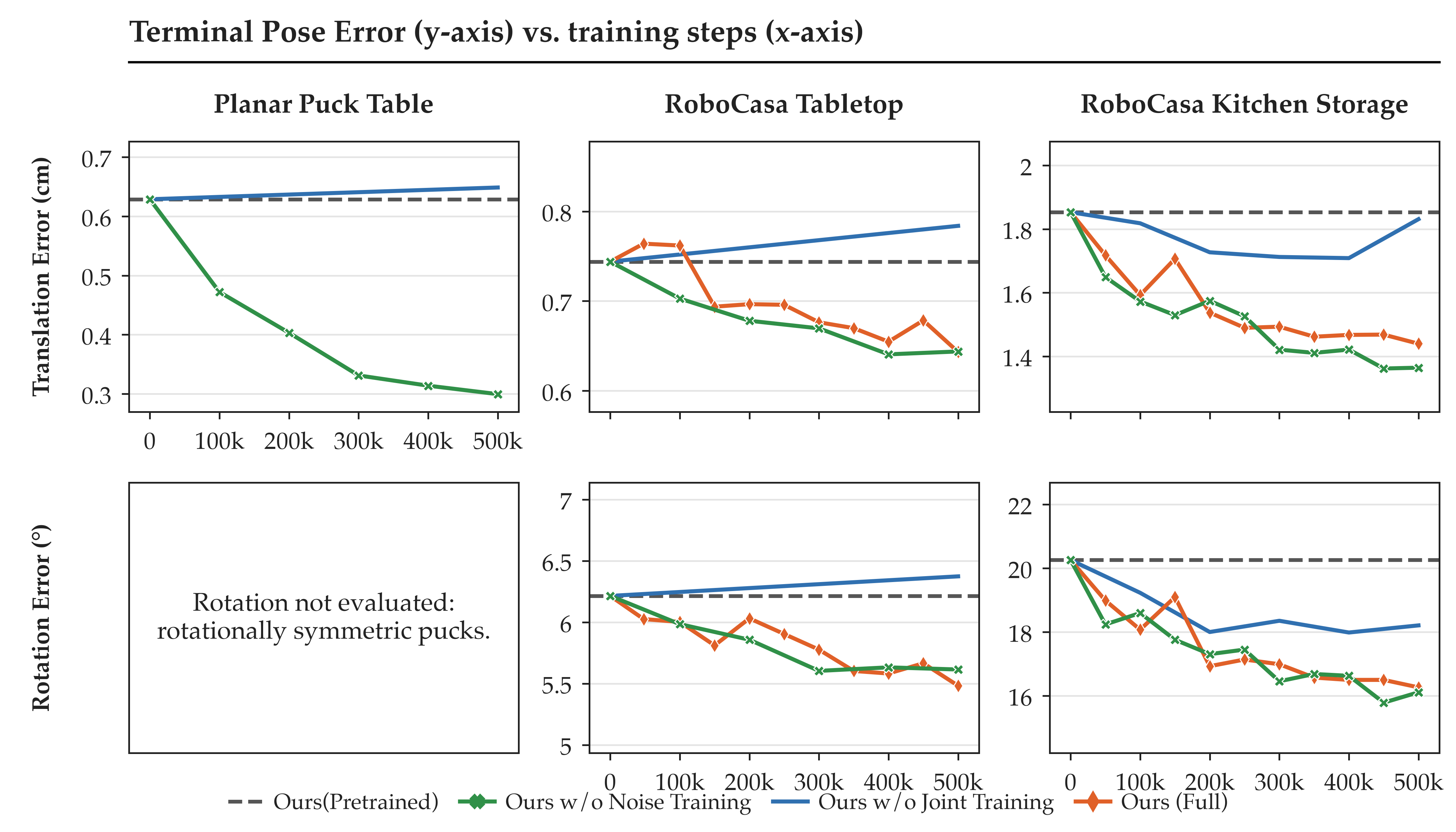}
    \caption{\textbf{Effect of joint training.} Terminal translation (top) and rotation (bottom) errors with shared pretrained initialization and matched additional optimization steps.}
    \label{fig:E2-fig-1}
    \vspace{-5mm}
\end{figure}

We evaluate long-horizon prediction and generalization to unseen objects against \textit{SlideSupport}~\cite{slide-support}, \textit{Coarse}~\cite{agboh-coarse-fine}, and \textit{PushEffect}~\cite{tekden2024pusheffect}. We additionally evaluate \textit{Ours(First-Contact)}, which restricts prediction to objects directly contacted by the target, \textit{Ours(Meshsubgraph)}, a mesh-subgraph ablation inspired by~\cite{fignet}, and \textit{Ours w/o Joint Training}, which trains our descriptor-based propagation network independently. Following~\cite{kipf2018neural}, we normalize errors by those of a static baseline that holds initial object poses fixed, yielding a dimensionless accuracy measure. All RoboCasa test objects are unseen during training. As shown in Fig.~\ref{fig:E1-fig-1}, DARE achieves the lowest normalized terminal errors across all evaluated horizons. We attribute this sustained advantage to the multi-step objective, which couples retrospective refinement with subsequent prediction to mitigate error accumulation. The advantage over \textit{Ours(First-Contact)} on puck and tabletop scenes highlights the importance of propagating disturbances beyond immediate target contacts. 

\begin{figure}[!t]
    \centering
    \includegraphics[width=1\linewidth]{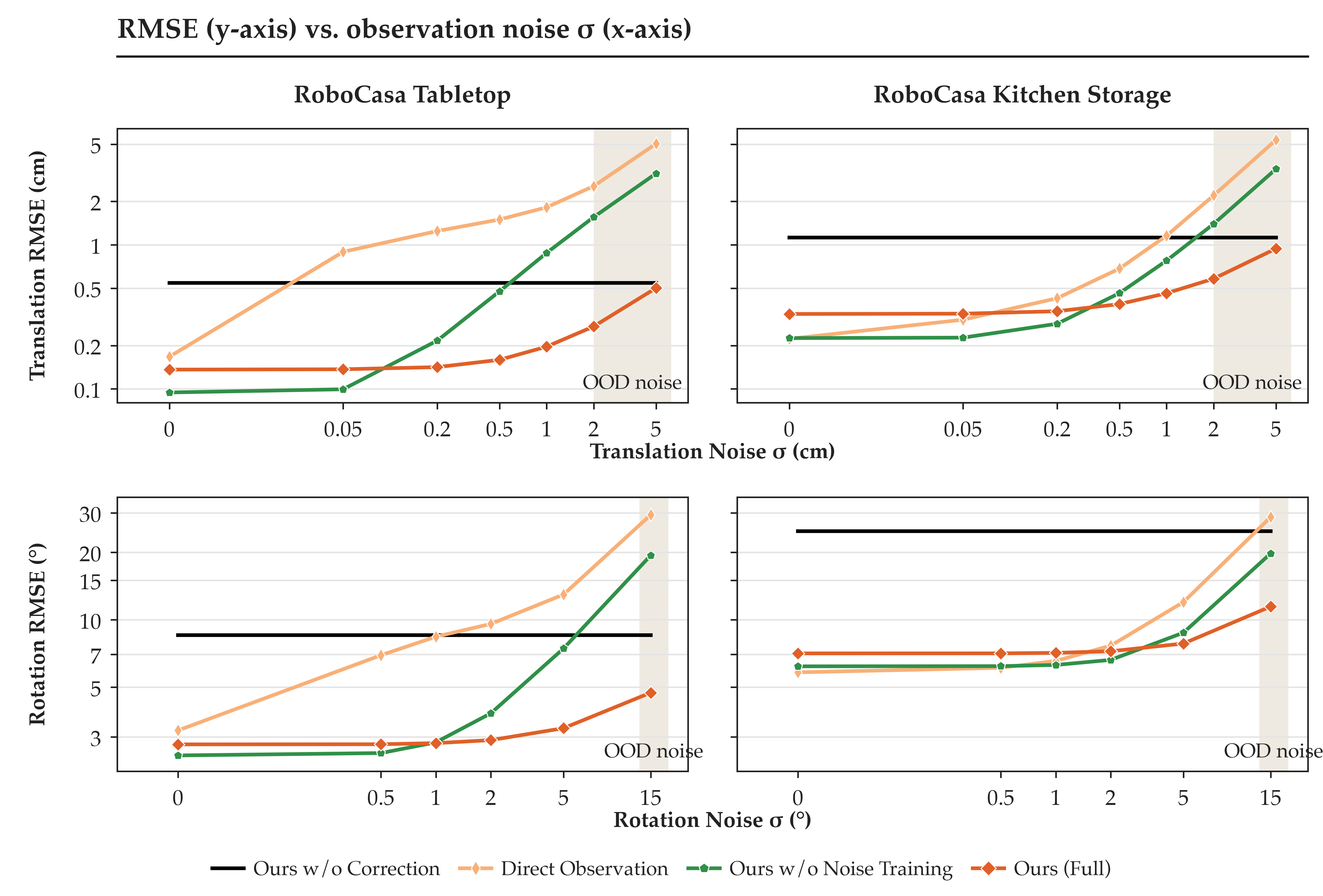}
    \caption{\textbf{Retrospective correction under noise.} Translation (top) and rotation (bottom) RMSE over 40-step rollouts versus observation noise $\sigma$, fixed within each rollout. Shading indicates noise levels beyond the noise training range.}
    \label{fig:E2-fig-2}
    \vspace{-5mm}
\end{figure}

\subsection{Retrospective Correction and Joint Training}
Starting from \textit{Ours(Pretrained)}, we compare continued propagation-only optimization (\textit{Ours w/o Joint Training}), joint predictor--corrector training with clean observations (\textit{Ours w/o Noise Training}), and joint training with noise augmentation (\textit{Ours (Full)}), using matched additional optimization steps and disabling correction during evaluation. For correction, we periodically perturb each bystander's observed pose with zero-mean Gaussian noise, varying its standard deviation across rollouts. Baselines include \textit{Ours w/o Correction}, which propagates predictions without observation updates, and \textit{Direct Observation}, which replaces predicted poses with observations. Fig.~\ref{fig:E2-fig-1} shows that joint training strengthens the predictor itself, indicating that retrospective supervision improves forward dynamics even without online correction. The sensitivity of direct observation updates to noise highlights the need to account for uncertainty when refining predicted states (Fig.~\ref{fig:E2-fig-2}). Training with clean observations supports correction under mild perturbations, while noise augmentation extends robustness to stronger and out-of-distribution noise, allowing observation updates to remain beneficial under uncertain feedback. Together, these findings connect joint training to improved dynamics learning and noise augmentation to reliable online refinement.

\subsection{Disturbance-aware Retrieval Planning}

We evaluate retrieval from cluttered cabinets containing 5, 10, or 20 objects, using 10 scenes per object count and difficulty level. Easy and hard scenes contain 20\% and 60\% tall objects, respectively, defined as those exceeding half the cabinet height. We compare against \textit{Direct retrieval}, which moves the target toward the goal, \textit{Collision avoidance}, which maintains 5\,mm clearance, and \textit{Ours w/o dist obj.}, which retains planning constraints but removes the disturbance objective. We evaluate under clean observations and under observations perturbed by zero-mean Gaussian noise with standard deviations of 2\,cm in position and $13^{\circ}$ in rotation. Under noise, we compare direct observation updates with noise-trained retrospective correction. Retrieval succeeds when the target reaches its goal without any bystander toppling or falling out of the cabinet. Translational, tilt, and yaw disturbances are computed separately by summing the corresponding initial-to-final pose changes over all bystanders remaining in the cabinet. Table~\ref{tab:e3-retrieval-combined} highlights the value of controlled contact: collision avoidance increasingly restricts retrieval as clutter grows, while direct retrieval can succeed at the cost of substantial collateral disturbance. Under clean observations, DARE maintains high retrieval success while generally reducing disturbance by balancing target progress against predicted contact consequences. These results suggest that long-horizon dynamics prediction supports safer contact-rich retrieval by anticipating indirect disturbance propagation and identifying motions that advance the target while limiting their effects on surrounding objects.

\subsection{Implementation}
All environments are built in MuJoCo~\cite{todorov2012mujoco} with RoboCasa objects and their ground-truth convex decompositions. The tabletop record has 18k training and 1k validation episodes; the cabinet record has 20k episodes of ten objects, 16.5k of them for training, and 1k unseen-object evaluation episodes. The propagation network uses six message-passing rounds, 256 hidden units, and a $0.35$\,m radius, pretrained on five-step autoregressive windows; the corrector is a two-layer 512-unit MLP. Joint training uses $W=C=5$ with a 2k-step corrector warm-up; observation noise is drawn per sample from $0.03$--$2$\,cm and $0.1$--$13^{\circ}$. The planner optimizes four-knot cubic splines over $H=10$ steps, executes $H_{\mathrm{e}}=5$, and runs ten rounds of 256 samples at temperature $0.05$. Rates are bounded by $0.03$\,m/s and $60^{\circ}$/s, tolerances are $\delta_p=3$\,cm and $\tau_{\max}=15^{\circ}$, and the cost uses $w_{\mathrm{term}}=2.5$, $w_p=1$, $w_{\tau}=0.172$, $w_{\psi}=0.0172$, and $d_{\max}=0.10$\,m, with the three disturbance weights halved in the 20-object scenes.

\section{Conclusion}

In this work, we introduced DARE, a relational dynamics framework for disturbance-aware object retrieval in clutter. Our experiments demonstrate that compact geometry descriptors provide an effective basis for predicting multi-object interactions across unseen shapes, connecting object geometry to scene-level physical consequences. Retrospective correction and joint training turn sparse observations into improvements in subsequent prediction, supporting long-horizon reasoning under noisy feedback. By integrating these predictions into closed-loop planning, DARE provides a foundation for robots to reason about and manage unavoidable contact when retrieving objects from clutters.

\bibliographystyle{IEEEtran} 
\bibliography{root}

\end{document}